\documentclass[sigplan,nonacm]{acmart}

\setcopyright{none}
\acmDOI{}
\acmISBN{}
\acmConference[Agent Skills '26]{The First Workshop on Agent Skills at ACM CAIS 2026}{May 26, 2026}{San Jose, CA, USA}
\acmBooktitle{Agent Skills '26: The First Workshop on Agent Skills at ACM CAIS 2026, May 26, 2026, San Jose, CA, USA}
\acmYear{2026}
\acmMonth{5}
\copyrightyear{2026}
\renewcommand\footnotetextcopyrightpermission[1]{}
\AtBeginDocument{%
}

\usepackage{adjustbox}
\renewcommand{\arraystretch}{1.15}
\usepackage{booktabs}
\usepackage{nicefrac}
\usepackage{microtype}
\usepackage{xcolor}
\usepackage{amsmath}
\usepackage{graphicx}
\usepackage{multirow}
\usepackage{tabularx}
\usepackage{array}
\usepackage{colortbl}
\usepackage{seqsplit}

\DeclareRobustCommand{\breakcode}[1]{\texorpdfstring{\texttt{\seqsplit{#1}}}{#1}}
\DeclareRobustCommand{\datapath}[1]{\breakcode{#1}}

\definecolor{helpgreen}{RGB}{34,139,34}
\definecolor{hurtred}{RGB}{178,34,34}
\definecolor{neutralgray}{RGB}{120,120,120}
\definecolor{lightblue}{RGB}{230,240,250}
\definecolor{lightyellow}{RGB}{255,255,230}

\title{FinSkillBench: Evaluating AI Agents and Domain Skills for Investment Management}

\author{Jermyn Zhen Yong Bek}
\authornote{Equal contribution.}
\email{jermynbek@gmail.com}
\affiliation{%
  \institution{Independent Researcher}
  \country{Singapore}}

\author{Zhuang Qiang Bok}
\authornotemark[1]
\email{zq.bok@deepinsightlabs.ai}
\affiliation{%
  \institution{Deep Insight Labs}
  \country{Singapore}}

\author{Zhongtian Sun}
\authornotemark[1]
\email{Z.Sun-256@kent.ac.uk}
\affiliation{%
  \institution{University of Kent}
  \city{Canterbury}
  \country{United Kingdom}}
\affiliation{%
  \institution{University of Cambridge}
  \city{Cambridge}
  \country{United Kingdom}}

\begin{document}

\begin{abstract}
Investment management is a high-stakes domain in which agentic AI systems must do more than generate plausible text. They must retrieve point-in-time data, assemble correct computational inputs, invoke specialized methods, and produce auditable structured outputs. We introduce \textbf{FinSkillBench}, an evaluation suite designed to measure whether language model agents can effectively use financial domain skills to solve investment management tasks. The benchmark spans three domains, portfolio construction, risk management, and fundamental analysis, and includes 12 subtasks with 2,603 task episodes. Each episode provides point-in-time inputs, hidden ground truth, and a task-specific verifier. We compare three conditions: \emph{no skill}, \emph{curated skill packages} consisting of procedural documents and executable components, and \emph{self-generated skills} in which the agent writes and reuses its own procedures within an episode. Across 9 models and a large-scale evaluation, curated skills consistently improve performance, raising mean scores from 0.366 to 0.528, with the largest gains in portfolio construction and risk management. In contrast, self-generated skills provide little benefit despite higher computational cost. An independent evaluation using a separate agent framework (Hermes Agent, 8 models, 5,280 episodes total) reproduces the directional pattern across all three domains, with the magnitude of skill effects varying by subtask and harness.

These results show that in investment management agents, access to reliable procedural skills can be as important as model choice, while naive self-generation of skills is often ineffective. We release the benchmark, evaluation tools, curated skill packages, and full trajectories to support further research.
\end{abstract}

\keywords{agent skills, LLM agents, benchmarking, evaluation, investment management, procedural knowledge}

\ccsdesc[500]{Computing methodologies~Machine learning}
\ccsdesc[300]{Information systems~Information retrieval}
\ccsdesc[300]{Information systems~Information extraction}

\maketitle

\section{Introduction}
 
Large language models are increasingly used as planning and analysis agents. In investment management, such agents are asked to solve tasks with quantitative structure: construct an optimal portfolio, monitor mandate violations, estimate stress losses, compute earnings-quality metrics, and normalize XBRL filings. These tasks differ from open-ended financial question answering. They require point-in-time discipline, structured outputs, reliable tool use, and verifiable numerical or symbolic correctness.
 
Existing evaluations often ask whether a model knows financial concepts or can answer finance questions. That is necessary but insufficient for evaluating agentic investment workflows. A useful investment agent must know when to load domain procedures, how to pass large numeric inputs without corrupting them, which formula applies to a filing metric, and how to return a result that can be audited by downstream software. The evaluation object is therefore not only the model, but the full process by which a model uses tools, documentation, and reusable skills.
 
We study this process through \textbf{FinSkillBench}, a benchmark and evaluation protocol for AI agents in investment management. The benchmark is organized around three domains: portfolio construction (mean-variance optimization, constrained optimization, tool-call parameterization, rebalancing, and Black–Litterman allocation), risk management (mandate monitoring, risk identification, stress testing, and remediation trades), and fundamental analysis (XBRL metric normalization, earnings quality, and driver decomposition).
 
The key intervention is \emph{skill availability}. In the \textbf{no-skill} condition, the agent receives the task without any tools and reusable domain documentation. In the \textbf{curated} condition, it can discover and load human-authored skill documents and tool scripts. In the \textbf{self-generated} condition, it must first write skill documents for itself and then use them. This design evaluates not just ``Can the model solve the task?'' but ``Does a reusable skill layer improve agentic financial analysis, and under what assumptions?''
 
\paragraph{Contributions.}
\begin{enumerate}
  \item A point-in-time investment-management benchmark with 2{,}603 task episodes across twelve subtasks and three domains.
  \item A factorial agentic protocol that isolates the effect of reusable skill access (no-skill, curated, self-generated) while holding tasks, tools, turn budget, and scoring fixed.
  \item Task-specific automated verifiers for numeric accuracy, constraint satisfaction, ranked risk recovery, attribution accuracy, trade matching, and financial-statement metric extraction.
  \item A full empirical run of \textbf{17{,}820 episodes} over 9 models showing that curated skills materially improve performance ($+16.2$~pp aggregate) while self-generated skills do not reliably do so in a one-episode setting ($+0.5$~pp).
  \item An independent \textbf{cross-harness validation} on a Hermes Agent with 8 models and 5{,}280  task episodes, which reproduces the pattern at the domain level and shows that the gains concentrate in specific subtasks.
  \item An analysis of failure modes and evaluation limitations for skill-based agent benchmarks, including scorer assumptions, benchmark heterogeneity, task-resource coupling, and the interpretation of skill gains.
\end{enumerate}
 
\section{Related Work}

\paragraph{Agent skills.} Anthropic introduced the Agent Skills specification as an open standard for equipping LLM agents with domain-specific procedural knowledge \citep{anthropic_skills2025}. SkillsBench \citep{skillsbench2026} provided a systematic evaluation across 86 tasks and 11 general domains, finding curated skills boost pass rates by approximately +16~pp on average while self-generated skills produce no average benefit. Our work extends this analysis to investment management.
 
\paragraph{Financial NLP benchmarks.} FinQA, ConvFinQA, and TAT-QA \citep{finqa2021,convfinqa2022,tatqa2021} evaluate numerical reasoning over financial reports and tables. Those datasets are valuable, but they primarily assess question answering over given evidence. FinSkillBench instead evaluates agentic workflows that combine point-in-time data retrieval, domain procedures, executable tools, and structured submission.
 
\paragraph{Tool-use and agent benchmarks.} AgentBench \citep{agentbench2023} and ToolBench \citep{toolbench2023} study multi-step tool interaction across broad domains; our benchmark narrows the domain and deepens the task contracts around investment-management methods. Classical financial methods such as mean-variance optimization \citep{markowitz1952}, Black–Litterman allocation \citep{bl1992}, factor exposure analysis \citep{ff1993}, Piotroski F-Score \citep{piotroski2000}, and Beneish M-Score \citep{beneish1999} define many of the ground-truth computations.
 
\section{Benchmark Design}
\label{sec:design}

\subsection{Domains and Subtasks}
 
FinSkillBench contains 12 subtasks (Appendix~\ref{app:subtasks}). Each task episode has a unique \texttt{task\_id}, a domain skill label, a subtask, an \texttt{as\_of\_date}, an input object shown to the agent, a hidden \texttt{expected\_output}, and a \texttt{verification} object specifying the scorer. Portfolio-construction and risk-management subtasks each expose 40 tasks per subtask in the full run; fundamental-analysis subtasks have larger filtered manifests (243 / 179 / 148 for normalization / earnings quality / driver decomposition) of which the runner uses the first 100 each.
 
\subsection{Data Stack and Point-in-Time Discipline}
 
Tasks are anchored in three data substrates. Portfolio construction and risk management use a \emph{market stack} (daily prices, Fama–French factors \citep{ff1993}, macroeconomic series) from which we derive a Ledoit–Wolf shrinkage covariance \citep{lw2004}, rolling factor exposures, a composite fundamentals-plus-momentum expected-return view, and a market-cap benchmark proxy. Fundamental analysis uses a \emph{reporting stack} built from SEC EDGAR submissions and XBRL CompanyFacts. Risk management additionally relies on \emph{constructed objects}---synthetic portfolios with planted risk profiles, mandate specifications, and stress-scenario definitions---which capture institutional middle-office inputs without conflating the task with filing parsing.
 
Three design choices govern the data flow. First, all episodes carry an explicit \texttt{as\_of\_date}; system prompts forbid reliance on post-date information and tooling returns only point-in-time slices. Second, task prompts apply \emph{lossy compaction} to large numeric structures (a covariance matrix is summarized in the visible prompt, while available at full precision through tool calls). This separation measures whether agents retrieve and pipe accurate numeric input rather than reasoning from rounded text. Third, the universe is a fixed 30-name sector-stratified sample S\&P 500, sacrificing breadth for controlled and reproducible evaluation.
 
\subsection{Ground Truth and Scorers}
 
The ground truth was generated offline, stored in each episode's \texttt{expected\_output}, and never shown to the agent. Labels fall into three regimes: \emph{exact numerical} (convex-solver outputs, Black–Litterman posteriors, stress P\&L); \emph{semi-structured ranked} (risk identification with partial credit for defensible alternative wordings); and \emph{accounting-derived} (canonical line items, screen components, segment-level bridges with tolerances). Each subtask is paired with a verifier whose structure matches the finance-meaningful object the task produces (Table~\ref{tab:scorers}). Full derivation pipelines and validation checks are provided in Appendix~\ref{app:data}, abbreviated input/output examples in Appendix~\ref{app:examples}, and complete scorer formulas in Table~\ref{tab:scorer_formulas}.
 
\begin{table*}[t]
\centering
\caption{Scorer-to-subtask mapping. Soft-vs-hard grading is calibrated to operational severity.}
\label{tab:scorers}
\footnotesize
\setlength{\tabcolsep}{5pt}
\renewcommand{\arraystretch}{1.15}

\begin{tabular}{@{}>{\centering\arraybackslash}p{2.7cm}>{\centering\arraybackslash}p{5.1cm}>{\centering\arraybackslash}p{4.9cm}@{}}
\toprule
\textbf{Subtask} & \textbf{Scorer} & \textbf{Object rewarded} \\
\midrule
Unconstrained optim. & \texttt{l2\_distance\_and\_objective} & Proximity to MV optimum \\
Constrained optim. &
\begin{tabular}[t]{@{}c@{}}
\texttt{constraint\_satisfaction\_and\_}\\
\texttt{objective}
\end{tabular}
& Constraint flags gate weight L2 \\
Tool-use param. & \texttt{parameter\_match} & Optimizer contract: nested constraint params \\
Rebalancing & \texttt{turnover\_compliance\_and\_objective} & Post-trade weights, turnover, trade-list consistency \\
Black--Litterman & \texttt{view\_specification\_and\_weights} & Posterior expected returns + posterior-optimal weights \\
Constraint monitoring & \texttt{exact\_match} & Mandate status + per-constraint accuracy \\
Risk identification & \texttt{ranked\_list\_recall} & Recovery + ordering of exposure categories \\
Stress testing & \texttt{absolute\_error} & Portfolio impact + sector/factor attribution \\
Risk remediation & \texttt{constraint\_satisfaction\_plus\_cost} & Feasible trades restoring compliance \\
Normalization & \texttt{metric\_absolute\_error} & Canonical line items within tolerance \\
Earnings quality & \texttt{earnings\_quality\_composite} & Piotroski components + Beneish flag + ratio terms \\
Driver decomposition & \texttt{driver\_f1\_and\_direction} & Named drivers, direction, magnitude \\
\bottomrule
\end{tabular}
\end{table*}

\section{Skills as a Scientific Manipulation}
\label{sec:protocol}

\subsection{Factorial Design Over Procedural Resources}
 
The central manipulation is \emph{which procedural resources an agent can access} while holding tasks, tools, turn budget, temperature, and scoring fixed. Three conditions define the factor:
\begin{itemize}
  \item \textbf{No-skill.} The skill-access interface is present but the skill corpus is empty: discovery calls resolve to nothing. Finance domain tools (e.g., \texttt{compute\_risk\_report}, \texttt{optimize\_portfolio}) were not available as well; the SKILL.md and scripts package is withheld. This is the reference baseline used for the main results in Section~\ref{sec:results}. The cross-harness validation in Section~\ref{sec:cross-harness} uses the same no-skill variant where both the SKILL.md package \emph{and} the domain tools are withheld; this design choice gave a fairer comparison, as discussed further in Section~\ref{aggregate}.
  \item \textbf{Curated.} Human-authored, domain specific skill and tool script packages were mounted. The agent can discover and read skills, and execute tools within the same episode. 
  \item \textbf{Self-generated.} Curated skill material is withheld; the agent receives write access to a per-episode scratch space and an explicit instruction to author one to three modular skill documents \emph{before} answering, then reload them.
\end{itemize}
The design is factorial over the \emph{resource}, not the \emph{interface}: tool signatures are invariant across conditions, isolating skill-content effects.
 
\subsection{What a Curated Skill Contains}
 
Each domain has exactly one curated package combining \textbf{declarative} content (Markdown procedures, formulas, common failure modes, API usage patterns) and \textbf{imperative} content (small validated Python entry points invoked through a constrained script-execution tool with JSON I/O). The split addresses a failure mode observed in piloting: language models reliably invent the shape of a computation but fail to pass large numeric structures through tool calls at full precision. A 900-float covariance matrix emitted by a model and re-parsed into an optimizer differs from the source at the last few digits, and those differences change the optimum. The curated design therefore keeps numeric-heavy routines inside validated scripts and provides a server-side mechanism to inject full-precision task fields into those scripts without round-tripping through model-generated JSON.
 
\subsection{Agent Loop}
 
The agent is a function-calling, ReAct-style loop with fixed system prompt, task prompt, and tool schema. Tools cover skill discovery and loading, skill authoring (self-generated only), retrieval of full task fields, script execution scoped to the active skill directory, domain data queries, and final JSON submission. Turn budget is capped at 12 with up to 3 retries for transient provider failures. Episodes that fail to emit valid JSON through the submission tool are scored \texttt{invalid\_submission} or \texttt{incomplete\_submission} rather than silently dropped. Verbatim prompts and tool availability by condition are listed in Appendix~\ref{app:prompts} (Table~\ref{tab:tools}).

\section{Experimental Setup}
\label{sec:setup}

\paragraph{Models.} We evaluated 9 models spanning frontier and open-weight families: \texttt{gpt-5.4}, \texttt{claude-sonnet-4.6}, \texttt{gpt-4.1}, \texttt{gemini-2.5-pro}, \texttt{grok-4}, \texttt{DeepSeek-V3.2}, \texttt{Phi-4}, \texttt{glm-5.1}, \texttt{gemini-3.1-flash-lite}.
 
\paragraph{Run.} Full coverage: 9 models $\times$ 3 conditions $\times$ 12 subtasks. Tasks per cell: 40 for PC and RM subtasks, 100 for FA subtasks. One sample per cell. Total: \textbf{17{,}820 episodes}. Execution: 12 concurrent OS processes (one per subtask runner), 16 \texttt{ThreadPoolExecutor} workers each. Temperature 0; max turns 12; eval retries 3.
 
\paragraph{Validity.} 17{,}176 / 17{,}820 evaluations (96.4\%) returned valid JSON. 644 (3.61\%) were \texttt{invalid\_submission}; 3{,}736 (21.0\%) hit the turn budget (\texttt{max\_turns\_exhausted}); 75 (0.4\%) were \texttt{incomplete\_submission}. \texttt{Phi-4} contributed 1{,}980 rows with mean score 0.000 (turn-budget exhaustion dominated). Most condition analyses exclude Phi-4 to avoid having a single non-functional agent dominate aggregates; the overall row reports both.

\section{Results}
\label{sec:results}

\subsection{Aggregate Skill Effect}
 

Pooled across all three conditions, the mean score over 17,820 evaluations is 0.375, or 0.422 once we drop Phi-4 (mean 0.000 across all 1,980 episodes). The per-condition means in Table~\ref{tab:aggregate} show that curated skills give a large, stable improvement over no-skill, while self-generated skills barely move the needle.
 
\begin{table}[ht]
\centering
\footnotesize
\caption{Aggregate condition means, excluding Phi-4 (5{,}280 evaluations per condition). CIs are bootstrap 95\% intervals (10{,}000 resamples) on paired deltas.}
\label{tab:aggregate}
\begin{tabular}{@{}lccc@{}}
\toprule
\textbf{Condition} & \textbf{Mean} & \textbf{$\Delta$ vs.\ No-Skill} & \textbf{95\% CI} \\
\midrule
No-Skill & 0.366 & --- & --- \\
\textbf{Curated} & \textbf{0.528} & \textcolor{helpgreen}{\textbf{+0.162}} & [+0.152, +0.171] \\
Self-Generated & 0.371 & \textcolor{neutralgray}{+0.005} & [$-$0.002, +0.011] \\
\bottomrule
\end{tabular}
\end{table}
 
\subsection{Domain-Level Effects}
 
Curated-skill gains are largest where the skill provides executable computational primitives (Table~\ref{tab:domain}).
 
\begin{table*}[t]
\centering
\small
\caption{Domain-level results (excl.\ Phi-4). PC and RM benefit most because curated skills contain validated solvers and precise input-passing instructions. CI shown for the curated $-$ no-skill paired delta.}
\label{tab:domain}
\begin{tabular}{@{}lccccc@{}}
\toprule
\textbf{Domain} & \textbf{No-Skill} & \textbf{Curated} & \textbf{Self-Gen} & \textbf{Curated $\Delta$} & \textbf{Self-Gen $\Delta$} \\
\midrule
Portfolio construction & 0.307 & \textbf{0.585} & 0.301 & \textcolor{helpgreen}{\textbf{+0.278}} [+0.258, +0.298] & $-$0.006 [$-$0.015, +0.004] \\
Risk management & 0.269 & \textbf{0.486} & 0.303 & \textcolor{helpgreen}{\textbf{+0.218}} [+0.195, +0.239] & +0.034 [+0.018, +0.049] \\
Fundamental analysis & 0.458 & \textbf{0.512} & 0.454 & \textcolor{helpgreen}{\textbf{+0.054}} [+0.045, +0.064] & $-$0.004 [$-$0.014, +0.006] \\
\bottomrule
\end{tabular}
\end{table*}
 
\subsection{Subtask-Level Effects}
 
The picture sharpens at subtask resolution (Table~\ref{tab:subtask}). The largest improvements occur on subtasks where the no-skill baseline is near zero or modest; subtasks with already-high baselines show small or negligible gains.
 
\begin{table*}[t]
\centering
\small
\caption{Subtask-level results (excl.\ Phi-4). Sorted by Curated $-$ No-Skill $\Delta$.}
\label{tab:subtask}
\begin{tabular}{@{}llcccc@{}}
\toprule
\textbf{Subtask} & \textbf{Domain} & \textbf{No-Skill} & \textbf{Curated} & \textbf{Self-Gen} & \textbf{$\Delta$} \\
\midrule
Unconstrained optimization & PC & 0.247 & 0.719 & 0.236 & \textcolor{helpgreen}{\textbf{+0.472}} \\
Constrained optimization & PC & 0.148 & 0.505 & 0.127 & \textcolor{helpgreen}{\textbf{+0.358}} \\
Risk identification & RM & 0.023 & 0.332 & 0.055 & \textcolor{helpgreen}{\textbf{+0.309}} \\
Rebalancing & PC & 0.183 & 0.451 & 0.164 & \textcolor{helpgreen}{+0.268} \\
Stress testing & RM & 0.017 & 0.281 & 0.027 & \textcolor{helpgreen}{+0.265} \\
Black--Litterman & PC & 0.153 & 0.375 & 0.147 & \textcolor{helpgreen}{+0.222} \\
Constraint monitoring & RM & 0.597 & 0.775 & 0.651 & \textcolor{helpgreen}{+0.178} \\
Risk remediation & RM & 0.438 & 0.558 & 0.478 & \textcolor{helpgreen}{+0.120} \\
Earnings quality & FA & 0.436 & 0.535 & 0.474 & \textcolor{helpgreen}{+0.099} \\
Tool-use parameterization & PC & 0.803 & 0.872 & 0.831 & \textcolor{helpgreen}{+0.069} \\
Driver decomposition & FA & 0.256 & 0.322 & 0.249 & \textcolor{helpgreen}{+0.066} \\
Normalization & FA & 0.681 & 0.678 & 0.638 & \textcolor{neutralgray}{$-$0.003} \\
\bottomrule
\end{tabular}
\end{table*}
 
The largest improvements occur in optimization and risk tasks where a correct answer depends on multi-step numerical computation. Tool-use parameterization has high no-skill performance because the expected output is a structured parameter translation rather than a full computation. Normalization shows no curated benefit because the base task is largely direct XBRL extraction plus simple arithmetic, already accessible via the global XBRL query tool.
 
\subsection{Model-Level Effects}
 
Skill effects are heterogeneous across base models (Table~\ref{tab:models}).
 
\begin{table*}[t]
\centering
\small
\caption{Per-model overall and condition means. ``Overall'' is the unweighted mean across all 12 subtasks averaged over the three conditions.}
\label{tab:models}
\begin{tabular}{@{}lccccc@{}}
\toprule
\textbf{Model} & \textbf{Overall} & \textbf{No-Skill} & \textbf{Curated} & \textbf{Self-Gen} & \textbf{Curated $\Delta$} \\
\midrule
gpt-4.1 & 0.488 & 0.330 & \textbf{0.735} & 0.398 & \textcolor{helpgreen}{\textbf{+0.405}} \\
gemini-2.5-pro & 0.490 & 0.380 & \textbf{0.678} & 0.412 & \textcolor{helpgreen}{\textbf{+0.298}} \\
DeepSeek-V3.2 & 0.521 & 0.458 & \textbf{0.658} & 0.447 & \textcolor{helpgreen}{+0.200} \\
gemini-3.1-flash-lite & 0.487 & 0.419 & \textbf{0.609} & 0.434 & \textcolor{helpgreen}{+0.191} \\
claude-sonnet-4.6 & 0.560 & 0.534 & \textbf{0.659} & 0.488 & \textcolor{helpgreen}{+0.125} \\
gpt-5.4 & 0.452 & 0.430 & \textbf{0.520} & 0.405 & \textcolor{helpgreen}{+0.091} \\
glm-5.1 & 0.268 & 0.274 & 0.285 & 0.245 & \textcolor{neutralgray}{+0.011} \\
grok-4 & 0.107 & 0.104 & 0.077 & \textbf{0.138} & \textcolor{hurtred}{$-$0.027} \\
Phi-4 & 0.000 & 0.000 & 0.000 & 0.000 & \textcolor{neutralgray}{0.000} \\
\bottomrule
\end{tabular}
\end{table*}
 
Curated skills do not simply help stronger models more. The largest gain accrues to \texttt{gpt-4.1} ($+$0.405), while \texttt{gpt-5.4} sees a smaller $+$0.091 gain and \texttt{claude-sonnet-4.6} starts from the strongest no-skill baseline (0.534) and gains $+$0.125. This suggests skill use is an interaction between model capability, tool-calling behavior, and skill-following reliability, not a monotone function of base model strength. Two failure modes prevent skill uptake entirely: \texttt{Phi-4} exhausts the turn budget on essentially every episode, and \texttt{grok-4} produces malformed final outputs at high rate.
 
\subsection{Cost and Interaction Overhead}
 
Self-generated skills are expensive in context and turns without delivering accuracy gains (Table~\ref{tab:cost}).
 
\begin{table*}[t]
\centering
\small
\caption{Per-episode resource use, excluding Phi-4.}
\label{tab:cost}
\begin{tabular}{@{}lcccc@{}}
\toprule
\textbf{Condition} & \textbf{Median Tokens} & \textbf{Mean Tokens} & \textbf{Median Latency (s)} & \textbf{Mean Turns} \\
\midrule
no-skill & 19{,}952 & 38{,}783 & 59.75 & 7.46 \\
curated & 22{,}546 & 35{,}378 & 49.79 & 7.30 \\
self-generated & 40{,}034 & 63{,}606 & 76.76 & 9.29 \\
\bottomrule
\end{tabular}
\end{table*}
 
Self-generated skills load in 97.7\% of non-Phi-4 episodes, but the generated skills rarely improve accuracy. ``Writing down a procedure'' inside the episode is not equivalent to having a validated domain skill with tested scripts and precise data-access guidance.

\section{Cross-Harness Validation}
\label{sec:cross-harness}

To test whether the no-skill vs.\ curated effect is robust to agent-implementation choices, we replicate the comparison on an independent harness: the Nous Research \textbf{Hermes Agent} \citep{nousresearch2026hermesagent} with the same fin domain skills, tools, prompt template, submission protocol, and turn budget (max\_turns~$=$~12). Only the Hermes \texttt{skills\_list} and \texttt{skill\_view} tools were activated, with all other Hermes built-in skills, toolsets, memory and personality turned off. Eight models---gpt-5.4, gpt-4.1, claude-sonnet-4.6, gemini-2.5-pro, glm-5.1, grok-4.20, DeepSeek-V3.2, gemma-4-31b-it---were each evaluated on 240 episodes (60 FA, 100 PC, 80 RM) across all the 12 subtasks , totaling \textbf{1{,}920} evaluations. Both runs were complete (zero runner errors) and used the same task data and scorers as the main study.
 
\subsection{Aggregate and Domain Effects}
\label{aggregate}

Table~\ref{tab:hermes} summarizes the Hermes replication at the aggregate, domain, and model levels.
\begin{table*}[t]
\centering
\small
\caption{Hermes harness: aggregate, by-domain, and by-model results. ``$\dagger$'' indicates the curated and no-skill 95\% CIs do not overlap. Numbers are unweighted means over 1,920 (overall, no-skill) and 5,280 (overall, curated skills) overall episodes. No-skill was a sample run of the first 20 episodes per each 12 subtask (240 total per model).}
\label{tab:hermes}
\begin{tabular}{@{}lcccccc@{}}
\toprule
\textbf{Group} & \textbf{$N$} & \textbf{No-Skill (95\% CI)} &  \textbf{$N$} &\textbf{Curated (95\% CI)} & \textbf{$\Delta$} & \\
\midrule
 &\multicolumn{6}{l}{\emph{Overall}} \\
All & 1920& 0.354 [0.338, 0.370]&  5280&0.679 [0.669, 0.688]& +0.325& $\dagger$ \\
\midrule
 &\multicolumn{6}{l}{\emph{By domain}} \\
Fundamental analysis & 480& 0.439 [0.408, 0.470]&  2400&0.567 [0.553, 0.582]	& +0.129& $\dagger$ \\
Portfolio construction & 800& 0.325 [0.301, 0.350]	&  1600&0.760 [0.744, 0.776]& +0.435& $\dagger$ \\
Risk management & 640& 0.325 [0.297, 0.354]&  1280&0.785 [0.767, 0.802]& +0.459& $\dagger$ \\
\midrule
 &\multicolumn{6}{l}{\emph{By model}} \\
claude-sonnet-4.6 & 240 & 0.304 [0.257, 0.353]&  660&0.795 [0.775, 0.815]& +0.490& $\dagger$ \\
DeepSeek-V3.2 & 240 & 0.371 [0.325, 0.419]& 660& 0.686 [0.659, 0.712]& +0.315&$\dagger$ \\
gemini-2.5-pro & 240 & 0.333	[0.291, 0.375]&  660&0.477	[0.441, 0.512]& +0.144& $\dagger$ \\
gemma-4-31b-it & 240 & 0.375	[0.332, 0.418]&  660&0.642	[0.614, 0.670]& +0.267& $\dagger$ \\
glm-5.1 & 240 & 0.397	[0.350, 0.444]&  660&0.779	[0.758, 0.801]	& +0.383& $\dagger$ \\
gpt-4.1 & 240 & 0.285	[0.244, 0.327]&  660&0.699	[0.672, 0.726]	& +0.414& $\dagger$ \\
gpt-5.4 & 240 & 0.408 [0.365, 0.451]&  660&0.660 [0.631, 0.687]& +0.252&$\dagger$ \\
grok-4.20& 240& 0.357 [0.314, 0.401]&  660&0.691 [0.664, 0.719]& +0.334& $\dagger$ \\
\end{tabular}
\end{table*}
 
The Hermes agent harness reproduces the directional pattern across all groups - at the overall level the no-skill baseline scored closely with the original harness (0.354 vs 0.366), and the Hermes harness had a higher uplift with curated skills (+0.325 vs +0.162). At the domain level: portfolio construction and risk management showed a substantial, statistically significant gain ($+$0.4 non-overlapping CIs), while fundamental analysis showed a smaller but significant change. Again, using the Hermes Agent provided a bigger uplift than the FinSkillBench harness. One factor could have drove this difference: the Hermes Agent harness load tools as Python functions, instead of reading scripts in the skills folder. This made tool calling and usage more efficient. 

\paragraph{Note on no-skill design across runs.}
Both the Section~\ref{sec:results}  and Section~\ref{sec:cross-harness} no-skill condition withholds domain tools, so the $+32.5$ pp aggregate Hermes uplift measures the same combined effect of skill documents and executable tool access. This is almost two times higher than the $+16.2$ pp curated skills uplift from the FinSkillBench agent harness. A clean decomposition of documents, tools, and their interaction would require a four-arm design: no-skill plus no-tools, no-skill plus tools, skill plus no-tools, and skill plus tools. We leave this to future work.
 
\subsection{Concentration of the Effect}
 
At the model level with the Hermes Agent harness, claude-sonnet-4.6 showed the biggest improvements wih skills (+0.49). Open weight models like GLM-5.1, DeepSeek-V3.2 and gemma-4-31b-it also showed good improvements of 38.3 pp, 31.5 pp and 26.7 pp respectively, with GLM-5.1 scoring the second highest (0.779) with curated skills. This proves that with appropriate skills and tools, local deployment of LLMs and financial agents yield good results. This is crucial in some financial settings for data privacy and security.  
 
\subsection{Cross-Harness Comparison: Same Pattern, Different Magnitude}
 
The Risk identification (domain: RM) subtask, one of the more difficult tasks, appears in both harnesses with similar baselines but larger skill gains in Hermes Agent: FinSkillBench harness 0.023~$\to$~0.332 ($+$0.309); Hermes harness 0.043~$\to$~0.457 ($+$0.414). Constrained optimization (domain: PC) in contrast, had similar baselines on the two harnesses (FinSkillBench: 0.148; Hermes: 0.177) and correspondingly different deltas ($+$0.358 vs.\ $+$0.087). The cross-harness data was consistent with one explanation that we leave to future work, larger studies to test rigorously: \emph{when the no-skill baseline is near zero, curated skills can produce large gains; when the no-skill baseline is already near the ceiling, skills cannot help further}. We report this as a hypothesis suggested by the data, not as a tested claim.
 
\subsection{Implications}
 
The cross-harness results have two practical implications. First, claims of ``$X$ pp skill uplift'' from any single benchmark are conditional on the no-skill baseline distribution implied by that harness's tools, prompts, and turn budget. Second, the signs of the cross-harness deltas are consistent (positive across all three domain; FA with the lowest improvement, PC and RM with higher improvements). A single benchmark and a single harness can therefore underestimate or overestimate skill effects depending on where its baselines sit. We recommend that future benchmarks report skill effects stratified by baseline level, not only as a single aggregate number.

\section{Discussion}
\subsection{What Curated Skills Measure}
 
Curated skill gains should not be interpreted as pure knowledge gains. In this benchmark, skills provide a combination of: (i) procedural reminders (which formula or workflow applies); (ii) API and tool-use guidance (correct argument names, parameter shapes); (iii) executable scripts implementing financial algorithms; and (iv) warnings about common failure modes (e.g., precision loss when passing covariance matrices through model-generated JSON). This combination reflects how practical agent systems are built. Production agents are rarely deployed as unaided language models; they are embedded in environments with documentation, tools, calculators, and guardrails. The benchmark therefore evaluates whether an agent can exploit such resources correctly.
 
\subsection{Why Self-Generated Skills Underperform}
 
The self-generated condition is intentionally strict. The agent must infer useful reusable knowledge, write it as a skill, reload it, and still solve the original task within 12 turns. This creates overhead and produces unvalidated instructions. The empirical result is that self-generated skills are usually not enough for one-shot improvement. They may still be useful in a multi-episode learning setting where generated skills persist, are tested, and can be corrected over time, but this benchmark does not evaluate that scenario.
 
\subsection{Evaluation as a Scientific Object}
 
The benchmark demonstrates that conclusions about ``financial agent capability'' depend strongly on evaluation design. For example: if invalid and max-turn submissions are dropped, score estimates are inflated relative to deployment-like reliability; if no-skill agents can call the same deterministic solvers as curated agents, the measured skill effect is confounded; if constrained-optimization is scored only by constraint-status labels, the metric may reward portfolios far from the optimizer solution; if Black–Litterman tasks expose fewer analyst views than the ground truth used, the task is unsolvable. These issues motivated the current scorer and runner design and illustrate why evaluation artifacts need documented assumptions, not just leaderboards.

\section{Limitations}
\textbf{Single-sample cells.} The full run uses one sample per (model, condition, task) cell. Our estimand is the population-level treatment effect of procedural-resource availability across the task distribution, not the stochastic success rate of any single trajectory. Because all conditions are evaluated on the same episodes with deterministic verifiers, we use paired comparisons, which absorb episode-level difficulty variance, and quantify uncertainty by bootstrap over paired deltas (10,000 resamples). We use temperature 0 to reduce within-cell variance, though this does not make provider-side execution fully deterministic. Repeated runs would tighten within-cell estimates but are not required to identify the population-level effect under our paired design. We acknowledge this leaves agent-loop variance unmeasured, particularly in turn-budget-exhausted episodes (21\% of calls), and flag it as a target for future expansion.
 
\textbf{Universe.} The 30-ticker S\&P 500 sample supports controlled evaluation but does not cover all asset classes, international markets, derivatives, private assets, or illiquid securities.
 
\textbf{Engine-derived labels.} Many ground-truth labels are produced by deterministic engines (convex solvers, Black–Litterman routines). The benchmark therefore evaluates agreement with those engines and their assumptions. Reasonable alternative modeling choices (e.g., shrinkage intensity) would yield slightly different labels.
 
\textbf{Driver decomposition ceiling.} Some expected driver labels depend on segment-level data not fully present in the processed XBRL panel. This caps attainable performance for agents that avoid unsupported domain guesses.
 
\textbf{Skill content includes scripts.} Curated skills include executable Python scripts. The evaluation is therefore about agent ability to discover and use domain skills, not about unaided financial reasoning---a feature for studying practical agents, but a caveat when comparing to text-only benchmarks.
 
\textbf{Time-dependent providers.} Tool-calling reliability, model versions, and API routing change. The artifact logs model identifiers and run configuration, but external reproducibility requires stable access to comparable model endpoints. \texttt{Phi-4} and \texttt{grok-4} in particular exhibited tool-calling failure modes on this run that may not generalize.
 
\textbf{Cross-harness scope.} While the Hermes replication had the same experimental scope and smaller sample episode size, it is sufficient to show that the directional skill effect on all domains transfers to a different harness. However, we should not establish equivalence of the two harnesses.
 
\section{Responsible Use and Ethics}
This benchmark is for evaluating analytical correctness in controlled investment-management tasks. It is not investment advice and should not be used to justify autonomous trading, suitability determinations, or client-specific recommendations. Financial agents can cause harm through erroneous calculations, hallucinated evidence, stale data, hidden assumptions, or inappropriate optimization objectives. The benchmark's point-in-time discipline, structured outputs, and explicit scoring are intended to make such failures easier to observe, not to eliminate them.
 
\section{Conclusion}
 
FinSkillBench evaluates investment-management agents as systems that combine language models, tools, task data, and reusable procedural knowledge. Across 17{,}820 full-run episodes over 9 models, curated skills produce a large and statistically stable improvement over no-skill baselines ($+$0.162; CI [+0.152, +0.171]), especially in optimization and risk workflows. Self-generated skills, by contrast, increase token and turn overhead without improving aggregate accuracy in this one-episode setting ($+$0.005; CI [$-$0.002, +0.011]). An independent cross-harness replication on Hermes Agent, covering 8 models, 1,920 no-skill episodes, and 5,280 curated episodes, reproduces the same domain-level pattern, with non-overlapping 95\% confidence intervals in every domain. The domain-level uplifts are $+12.9$ pp for FA, $+43.5$ pp for PC, and $+45.9$ pp for RM. The Hermes uplift is larger in absolute terms than the FinSkillBench-Skills uplift, partly because the no-skill condition in this cross-harness setting withholds domain tools as well as the \texttt{SKILL.md} package, as discussed in Section~\ref{aggregate}. These findings argue for evaluating domain agents under explicit resource conditions and for treating skills, tools, and scoring assumptions---and the harness itself---as first-class objects of evaluation.

\begin{acks}
We thank the Agent Skills '26 reviewers for their feedback. We release FinSkillBench task episodes, curated skill packages, evaluation harnesses, and full agent trajectories to support reproducibility.
\end{acks}

\balance
\bibliographystyle{ACM-Reference-Format}
\bibliography{references}

\appendix
\begingroup\sloppy

\section{Data Processing and Ground Truth Derivation}
\label{app:data}

\subsection{Universe and Shared Infrastructure}

All three domains share a common 30-ticker universe drawn from the S\&P~500, stratified across all 11 GICS sectors. The universe is frozen at a single snapshot date to avoid survivorship bias. Sector distribution: Technology~(5), Financial Services~(4), Industrials~(4), Healthcare~(3), Consumer Cyclical~(3), Consumer Defensive~(2), Energy~(2), Real Estate~(2), Basic Materials~(2), Utilities~(2), Communication Services~(1). Each record carries symbol, CIK, ISIN, CUSIP, sector, sub-sector, exchange, country, and S\&P~500 addition date.

All three data pipelines follow the same architecture:
\begin{center}
\texttt{Download} $\to$ \texttt{Process} $\to$ \texttt{Construct} $\to$ \texttt{Ground Truth} $\to$ \texttt{Tasks} $\to$ \texttt{Validate}
\end{center}
Each pipeline is orchestrated by a single \texttt{run\_pipeline.py} that loads the security master, creates a shared context, resolves which phases to run, and executes them in dependency order.

\subsection{Portfolio Construction Data}
\label{app:data:pc}

\paragraph{Raw data sources.}
\begin{itemize}
  \item \textbf{Daily Prices}: Daily OHLCV records per ticker. Each record contains \texttt{date}, \texttt{open}, \texttt{high}, \texttt{low}, \texttt{close} (split-adjusted), \texttt{adjClose}, and \texttt{volume}. 30~files, covering 2024-01 through 2025-12.
  \item \textbf{Dividend-Adjusted Prices}: Separate adjusted-close series accounting for dividends and splits.
  \item \textbf{Historical Market Capitalization}: Daily market cap per ticker, used to construct benchmark weights.
  \item \textbf{Fama--French Factors} \citep{ff1993}: Daily returns for the 5-factor model (Mkt-RF, SMB, HML, RMW, CMA) plus momentum.
  \item \textbf{Macroeconomic Series}: 8~time series---10-Year Treasury, 2-Year Treasury, 3-Month Treasury, yield curve spread, CPI, unemployment rate, Fed Funds rate, and VIX.
\end{itemize}

\paragraph{Processing logic.}
\begin{enumerate}
  \item \textbf{Price panel}: \datapath{processed/prices.parquet}. Per-ticker JSON files are merged into a wide-format panel (15{,}060 rows $\times$ 9~columns, 24~months, 30~symbols). Missing dates are forward-filled; tickers with fewer than 120~observations are flagged.
  \item \textbf{Returns}: \datapath{processed/returns.parquet}. Daily simple returns computed as $(P_t - P_{t-1})/P_{t-1}$ from adjusted close prices.
  \item \textbf{Factor exposures}: \datapath{processed/factor\_exposures.parquet}. For each ticker and each month-end date, a rolling 252-day OLS regression of excess returns against the 6~factors (FF5~+~momentum) produces per-ticker factor betas. The regression requires a minimum of 120~observations.
  \item \textbf{Benchmark weights}: \datapath{processed/benchmark\_weights.parquet} (240~rows). Monthly market-cap-weighted benchmark across the 30~symbols. For each month-end, weights are proportional to market capitalization: $w_i = \text{mcap}_i / \sum_j \text{mcap}_j$.
  \item \textbf{Covariance matrices}: \datapath{processed/covariance/\{date\}.npz} (12~files). Ledoit--Wolf shrinkage covariance \citep{lw2004} estimated from a 252-day trailing window of daily returns, annualized by multiplying by 252. Stored as compressed NumPy archives with \texttt{cov} (30$\times$30 array) and \texttt{symbols} (ticker ordering). Positive semi-definiteness is verified; if the minimum eigenvalue is negative, a ridge correction $\Sigma \leftarrow \Sigma + (|\lambda_{\min}| + 10^{-8})I$ is applied.
  \item \textbf{Expected-return views}: \datapath{processed/views/\{date\}.json} (22~files). A composite signal from 5~alpha factors, each cross-sectionally z-scored and combined with fixed weights: earnings yield (25\%), ROE (20\%), 12-1 momentum (25\%), EPS revision (15\%), low-volatility (15\%). The composite z-score is scaled to daily expected returns via Grinold--Kahn alpha methodology. Each file contains per-ticker signal values, composite z-scores, and the resulting expected returns.
  \item \textbf{Constraint sets}: \datapath{processed/constraints/\{easy,medium,hard\}.json}. Three tiers of increasing restrictiveness:
  \begin{itemize}
    \item \texttt{easy}: long-only, max 6\%/name, max 25\%/sector, 20--30 names.
    \item \texttt{medium}: long-only, max 6.6\%/name, max 20\%/sector, 18--30 names, turnover $\leq$ 20\%, tracking error $\leq$ 3\%.
    \item \texttt{hard}: long-only, max 7.7\%/name, max 15\%/sector, 15--30 names, turnover $\leq$ 15\%, tracking error $\leq$ 2\%, factor-neutral on Mkt-RF, SMB, HML.
  \end{itemize}
\end{enumerate}

\paragraph{Ground truth derivation.} For each (date, difficulty, objective) triple where both covariance and views exist, we run \texttt{optimize\_portfolio()} via \texttt{cvxpy}---a convex quadratic program with the specified objective and constraints. The solver tries SCS first (handles SOCP/conic programs), falling back to OSQP (QP only). The four objectives are formulated as:
\begin{itemize}
  \item \texttt{max\_sharpe}: maximize $\mu^\top w - r_f - \lambda \cdot w^\top \Sigma w$ (scalarized utility).
  \item \texttt{min\_variance}: minimize $w^\top \Sigma w$.
  \item \texttt{max\_return}: maximize $\mu^\top w$.
  \item \texttt{risk\_parity}: minimize $\|w - w_{\text{inv-var}}\|_2^2 + 0.5 \cdot w^\top \Sigma w$, where $w_{\text{inv-var}}$ are inverse-variance weights.
\end{itemize}
This produces 144~main ground truth files (12~dates $\times$ 3~difficulties $\times$ 4~objectives). Of these, 96 are optimal and 48 are infeasible (all hard-difficulty, where factor neutrality plus tight constraints are infeasible with 30~symbols).

For rebalancing ground truth, a two-stage procedure is used: Stage~1 solves the standard optimization; Stage~2 minimizes one-way turnover $\|w - w_0\|_1 / 2$ subject to staying within a tight tolerance of the Stage~1 objective value.

For Black--Litterman, for each of 12~dates $\times$ 4~view variants (different random seeds controlling number of views, symbols, return magnitudes, confidence levels), we run the BL model \citep{bl1992}. The model computes equilibrium returns $\pi = \delta \Sigma w_{\text{mkt}}$ from market-cap weights, constructs the pick matrix $P$ and view vector $Q$ from analyst views, and computes the posterior via $\mu_{\text{BL}} = [(\tau\Sigma)^{-1} + P^\top \Omega^{-1} P]^{-1} [(\tau\Sigma)^{-1}\pi + P^\top \Omega^{-1} Q]$ where $\Omega_{kk} = (1 - c_k) \cdot P_k \Sigma P_k^\top \cdot \tau$ and $c_k$ is the confidence of view $k$. This produces 48~BL ground truth files.

\paragraph{Example: ground truth for \breakcode{2025-01-31\_easy\_max\_sharpe}.}
The optimizer produces weights across 30~tickers: AIG~6.0\%, BF-B~6.0\%, BR~6.0\%, CMG~6.0\%, CTRA~6.0\%, DHR~6.0\%, KHC~6.0\%, L~6.0\%, LHX~6.0\%, LNT~6.0\%, REG~6.0\%, TTWO~6.0\% (all at the 6\% cap), with smaller allocations to BAX~1.3\%, CPRT~4.8\%, DD~2.5\%, HUM~2.2\%, LYV~0.6\%, SBUX~3.5\%, SRE~4.7\%, TT~3.5\%, ZBRA~4.8\%, and zero weight on C, CARR, CPAY, EFX, EQT, ON, USB, VMC. Expected return: 0.077\%/day; risk: 10.76\%; Sharpe: 0.0071. All constraints satisfied.

\paragraph{Ground truth validation.} All 320~PC and RM ground truth entries were independently validated through two layers: (i)~\emph{property checks} verifying structural invariants (weight sums, non-negativity, constraint satisfaction, rank ordering, trade arithmetic), and (ii)~\emph{re-derivation} re-running the same computation from stored inputs and comparing against stored expected outputs. Re-derivation confirms exact reproducibility for unconstrained optimization, constrained optimization, Black--Litterman, and stress testing (all L2~$<$~0.001 or absolute error~$<$~0.005). Zero errors, zero warnings across all 320~tasks.

\paragraph{Ground truth patches.} Two bugs were discovered and fixed before the evaluation run. (i)~The original pipeline generated ``unconstrained'' GT using the \texttt{easy} constraint set; a patch re-ran the optimizer with truly unconstrained settings (\texttt{long\_only=True} only). (ii)~The \texttt{black\_litterman()} function overwrote \texttt{view\_adjustments} each loop iteration instead of appending; a patch fixed the bug, reconstructed original views from the same RNG seed, and re-ran the BL model. All original files were backed up.

\subsection{Risk Management Data}
\label{app:data:rm}

\paragraph{Raw data sources.}
\begin{itemize}
  \item \textbf{Daily Prices}: Shared price data from the PC pipeline (30~tickers, 2017--2025).
  \item \textbf{Fama--French Factors}: Daily factor returns for FF5~+~momentum.
  \item \textbf{Macroeconomic Series}: 9~series---Fed Funds Effective Rate, VIX, 2-Year Treasury, 10-Year Treasury, yield curve spread, high-yield credit spread, CPI, unemployment rate, and trade-weighted USD index. Daily series are used directly; monthly series (CPI, unemployment) are forward-filled to daily frequency.
\end{itemize}

\paragraph{Processing logic.}
\begin{enumerate}
  \item \textbf{Returns}: \datapath{processed/returns/}. Daily simple returns, log returns, 21-day rolling annualized volatility ($\sigma_i = \text{std}(r_{i,t-20:t}) \times \sqrt{252}$), and running max drawdown from trailing peak.
  \item \textbf{Factor exposures}: \datapath{processed/factor\_exposures.parquet}. Rolling 252-day OLS regression of each ticker's excess returns against the 6~factors. Minimum 120~observations required. Produces per-ticker, per-date betas: $\beta_{\text{mkt\_rf}}, \beta_{\text{smb}}, \beta_{\text{hml}}, \beta_{\text{rmw}}, \beta_{\text{cma}}, \beta_{\text{mom}}$. Median absolute beta across all estimates: 0.429; range: [$-$2.55, $+$4.52].
  \item \textbf{Covariance}: \datapath{processed/covariance/cov\_\{date\}.npz}. Monthly Ledoit--Wolf shrinkage covariance, same methodology as PC. All 12~matrices verified PSD.
  \item \textbf{Macro panel}: \datapath{processed/macro\_panel.parquet}. Merged macro indicators, forward-filled to daily frequency for point-in-time access.
\end{enumerate}

\paragraph{Constructed data.} This is the RM-specific step that creates synthetic evaluation objects with deliberately planted risk characteristics.

\paragraph{Portfolios (200~files, 5~types $\times$ 40~each).}
\begin{itemize}
  \item \emph{Concentrated}: 4~names at 15\% each (60\% total in one sector), remaining 15~names at $\sim$2.7\% each. Planted risks: \texttt{sector\_concentration}, \texttt{single\_name\_risk}, \texttt{low\_diversification}. Example: AIG, C, L, USB each at 15\% (Financial Services sector), plus 15~diversifying names.
  \item \emph{Factor-tilted}: Overweight high-beta or high-momentum names to create extreme factor exposures ($|\beta| > 0.5$ on at least one factor).
  \item \emph{Liquidity-stressed}: Overweight small-cap or low-volume names from the universe.
  \item \emph{Mandate-violating}: Deliberately violates common mandate constraints (position limits, sector limits, beta range).
  \item \emph{Stress-vulnerable}: High sensitivity to specific stress scenarios (e.g., concentrated in rate-sensitive sectors for interest-rate shocks).
\end{itemize}

\paragraph{Mandates (12~files across 3~difficulty tiers).}
\begin{itemize}
  \item \emph{Easy} (2~mandates): Simple Long-Only (max 8\%/name, max 30\%/sector); Diversified Equity (max 5\%/name, min 50bp, max 25\%/sector).
  \item \emph{Medium} (3~mandates): Balanced with tracking error $<$ 3\%, beta 0.8--1.2, turnover $<$ 15\%; Conservative with max 4\%/name, max 20\%/sector, TE $<$ 2\%, beta 0.7--1.0, HHI $<$ 0.08.
  \item \emph{Hard} (3~mandates): Institutional Core (max 4\%/name, TE $<$ 2.5\%, beta 0.9--1.1, turnover $<$ 10\%, market beta $<$ 1.1); Enhanced Index (max 3.5\%/name, TE $<$ 1.5\%, beta 0.95--1.05, turnover $<$ 8\%, HHI $<$ 0.05).
\end{itemize}

\paragraph{Stress scenarios (12~files: 4~historical replays + 8~hypothetical).}
\begin{itemize}
  \item \emph{Historical replays}: COVID shock (Feb--Apr 2020), 2018 selloff (Oct--Dec 2018), 2022 tightening (Jan--Oct 2022), 2023--24 rebound.
  \item \emph{Hypothetical (rates/credit/equity)}: Rates $+$200bp (equity $-$5\%, credit $+$50bp); Rates $-$100bp (equity $+$2\%, credit $-$20bp); Equity crash $-$20\% (credit $+$200bp); Credit crisis (equity $-$10\%, credit $+$300bp).
  \item \emph{Hypothetical (macro/rotation)}: Correlation spike (target $\rho = 0.8$); Sector rotation (tech $-$15\%, value $+$5\%); Strong dollar (USD $+$10\%, EM equity $-$8\%); Stagflation (CPI $+$3\%, GDP $-$2\%, rates $+$100bp).
\end{itemize}
Each scenario specifies factor-level shocks and per-sector sensitivities.

\paragraph{Ground truth derivation.}
All RM ground truth is deterministic---computed from the constructed portfolios, mandates, scenarios, and processed market data using a reference risk engine (\breakcode{\_risk\_engine.py}).
\begin{itemize}
  \item \textbf{Risk identification} (200~GT): For each portfolio, the engine computes concentration metrics (HHI $= \sum w_i^2$, top-5 weight share), sector exposures, portfolio-level factor betas ($\beta_p = \sum w_i \beta_i$), and VaR/CVaR at 95\% confidence from the trailing 252-day return distribution. Risks are ranked by magnitude into types: \texttt{factor\_tilt}, \texttt{sector\_concentration}, \texttt{single\_name}, \texttt{concentration}, and \texttt{liquidity\_risk}.
  \item \textbf{Constraint monitoring} (600~GT): For each (portfolio, mandate) pair, each constraint is evaluated: position limits, sector limits, beta range, long-only ($w_i \geq 0$), and turnover (if applicable). Each constraint produces a PASS/FAIL status with the worst offender.
  \item \textbf{Stress testing} (2{,}400~GT): For each (portfolio, scenario) pair, the engine applies factor-level shocks to the portfolio's factor exposures,
  \[
    \text{PnL}_i = \sum_f \beta_{i,f} \cdot \text{shock}_f + \text{sector\_sensitivity}_i,\quad
    \text{PnL}_p = \sum_i w_i \cdot \text{PnL}_i.
  \]
  Attribution decomposes total P\&L into sector and factor contributions. All 40~stress-testing GT entries reproduce to within 0.005 absolute error on re-derivation.
  \item \textbf{Risk remediation} (400~GT): For each non-compliant (portfolio, mandate) pair, a constrained optimizer generates minimum-turnover trades to restore compliance. Trade arithmetic is verified via $w_{\text{current}} + \Delta w = w_{\text{target}}$ for every trade (tolerance $10^{-4}$).
\end{itemize}

\paragraph{Example: constructed portfolio.}
\breakcode{portfolio\_concentrated\_000}: 4~Financial Services names (AIG, C, L, USB) each at 15.0\%, plus 15~diversifying names at 2.7\% each. Total NAV: \$100M. Planted risks: sector concentration (Financial Services at 60\%), single-name risk (4~names $>$ 10\%), low diversification (HHI $\approx$ 0.10).

\paragraph{Example: stress scenario.}
\breakcode{stress\_rates\_down\_100}: Hypothetical flight-to-safety scenario. Shocks: interest rates $-$100bp (parallel shift), equity market $+$2\% (correlated move), credit spread $-$20bp. Sector sensitivities: Real Estate $-$15\%, Utilities $-$10\%, Technology $-$8\%, Healthcare $-$3\%, Financial Services $+$3\%.

\paragraph{Example: mandate.}
\breakcode{mandate\_enhanced\_index} (hard): 8~constraints---position limit max 3.5\%, sector limit max 20\%, tracking error $<$ 1.5\%, beta 0.95--1.05, monthly turnover $<$ 8\%, gross exposure $\leq$ 100\%, long-only, HHI $<$ 0.05.

Task episodes are stratified samples of 40~tasks per subtask from the GT pool, balanced across portfolio types, regimes, and difficulty levels (160~RM tasks total).

\subsection{Fundamental Analysis Data}
\label{app:data:fa}

\paragraph{Raw data sources.}
\begin{itemize}
  \item \textbf{Regulatory Filing Metadata}: Filing metadata per company (30~files). Contains filing history, company info, and identifier mappings.
  \item \textbf{XBRL CompanyFacts}: All reported XBRL facts for a company across all filings, organized by taxonomy and tag (30~files). This is the primary structured data source for financial metrics.
  \item \textbf{Pre-Parsed Financial Statements}: Quarterly and annual income statements, balance sheets, cash flow statements, key metrics, and financial ratios. Used as cross-reference for ground truth validation.
  \item \textbf{Revenue Segmentation}: Revenue breakdown by product/business segment and by geographic region. Used for driver decomposition ground truth.
  \item \textbf{Parsed Filing Sections}: Extracted sections from 10-K/10-Q filings---MD\&A, financials, footnotes, and semantic tree. Coverage: 93\% for MD\&A, 17\% for footnotes.
\end{itemize}

\paragraph{Processing logic.}
\begin{enumerate}
  \item \textbf{Filing timeline}: \datapath{processed/filing\_timeline.parquet}. Merges regulatory filing metadata with pre-parsed financial statement period data. Maps each (symbol, period\_end) to form type (10-K, 10-Q, 20-F, 40-F), filing date, accession number, filer type (domestic/foreign), and XBRL taxonomy (us-gaap/ifrs-full). 287~total filings, 281~cross-matched (98\%).
  \item \textbf{XBRL panel}: \datapath{processed/xbrl\_panel.parquet} (2{,}880~rows $\times$ 17~columns). Extracts canonical financial metrics from CompanyFacts JSON. XBRL tag names are normalized via an alias mapping to canonical metric names. Example revenue tags include \texttt{Revenues}, \breakcode{RevenueFromContractWithCustomerExcludingAssessedTax}, and \texttt{SalesRevenueNet}, all mapped to \texttt{revenue}. Each row contains: symbol, period\_end, canonical metric name, value, duration class (\texttt{instant} for balance-sheet items; \texttt{quarterly}, \texttt{ytd}, or \texttt{annual} for flow items), and filing metadata. Coverage: 30~symbols $\times$ 46~unique periods.
  \item \textbf{YTD-to-quarterly conversion}: For flow metrics (revenue, net income, operating cash flow, etc.) where only year-to-date values are available, quarterly values are derived by differencing consecutive YTD entries: $Q_t = \text{YTD}_t - \text{YTD}_{t-1}$. The pipeline uses actual period lookups from the XBRL panel rather than fixed calendar offsets, handling fiscal-year misalignment correctly.
  \item \textbf{Computed fallbacks}: When \texttt{total\_liabilities} is not directly reported in XBRL, it is computed as $\text{total\_assets} - \text{stockholders\_equity}$. Stock metrics (balance sheet) prefer \texttt{instant} duration class; flow metrics prefer \texttt{quarterly}, then derived from YTD.
\end{enumerate}

\paragraph{Ground truth derivation.} All FA ground truth is computed from raw XBRL and pre-parsed financial statement data---no optimizer or model involved. Cross-referencing between the two sources provides validation (75.4\% exact agreement rate).

\medskip
\noindent\textbf{Normalization} (244~GT): For each (ticker, period\_end), 10~canonical metrics are extracted and cross-referenced:
\begin{itemize}
  \item Direct from XBRL and pre-parsed statements: revenue, operating\_income, net\_income, eps\_diluted, total\_assets, total\_liabilities, stockholders\_equity, operating\_cash\_flow.
  \item Computed: operating\_margin $=$ operating\_income / revenue; ebitda $=$ operating\_income $+$ depreciation\_amortization.
  \item Each metric carries a confidence level: \texttt{high} (both structured sources agree within 1\%), \texttt{medium} (computed/derived), \texttt{low} (single-source flow metric).
\end{itemize}

\medskip
\noindent\textbf{Earnings quality} (1{,}002~GT): For each (ticker, period\_end), the pipeline computes from quarterly financial statements:
\begin{itemize}
  \item \emph{Piotroski F-Score} \citep{piotroski2000}: 9~binary components testing profitability (ROA positive, CFO positive, $\Delta$ROA positive, accruals negative), leverage ($\Delta$leverage negative, $\Delta$current ratio positive, no dilution), and efficiency ($\Delta$gross margin positive, $\Delta$asset turnover positive). Each component requires current and prior-year-same-quarter data.
  \item \emph{Beneish M-Score} \citep{beneish1999}: 8~ratio components (DSRI, GMI, AQI, SGI, DEPI, SGAI, LVGI, TATA) combined into a composite score. $M > -1.78$ flags manipulation risk.
  \item \emph{Accruals ratio}: (net\_income $-$ operating\_cash\_flow) / total\_assets.
  \item \emph{Income quality ratio}: operating\_cash\_flow / net\_income.
  \item \emph{Flags}: Triggered by low Piotroski ($\leq 3$), Beneish manipulation flag, high accruals ($> 0.05$), or negative income quality.
\end{itemize}

\medskip
\noindent\textbf{Driver decomposition} (997~GT): For each consecutive quarter pair, the pipeline computes:
\begin{itemize}
  \item Revenue delta: $\Delta R = R_{\text{current}} - R_{\text{prior}}$.
  \item Top revenue drivers: Decomposed into product-segment and geographic-segment contributions from revenue segmentation data. Each driver has a name, type (\texttt{product\_segment} or \texttt{geographic\_segment}), direction (increase/decrease), magnitude in USD, and contribution percentage.
  \item Margin drivers: Changes in operating margin, gross margin, SGA ratio, R\&D ratio, tax ratio, COGS ratio, D\&A ratio, and interest ratio between the two periods, each with direction and delta in percentage points.
\end{itemize}

\paragraph{Example: normalization ground truth for TRV (2024-12-31, 10-K).}
Revenue: \$46.4B (high confidence); net income: \$5.0B (high); EPS diluted: \$21.47 (high); total assets: \$133.2B (high); stockholders' equity: \$27.9B (high); operating cash flow: \$9.1B (high); operating margin: 13.31\% (computed); EBITDA: \$6.9B (computed, medium confidence). Cross-source agreement: true.

\paragraph{Example: earnings quality ground truth for AAPL (2022-09-24, 10-Q).}
Piotroski score: 8/9 (strong). Components: all positive except \texttt{delta\_gross\_margin\_positive}~$= 0$. Beneish M-score: $-2.31$ (no manipulation flag). Component detail: DSRI~1.19, GMI~1.02, AQI~0.91, SGI~1.09, DEPI~1.02, SGAI~0.99, LVGI~1.06, TATA~$-0.01$. Accruals ratio: $-0.0097$ (high quality). Income quality ratio: 1.16. Flags: none.

\paragraph{Example: driver decomposition ground truth for AAPL (Q2$\to$Q3 2017).}
Revenue delta: $-$\$7.5B. Top 5 geographic drivers: Greater China $-$\$2.7B (36.4\%), Europe $-$\$2.1B (27.5\%), Rest of Asia Pacific $-$\$1.1B (14.2\%), Japan $-$\$0.9B (11.5\%), Americas $-$\$0.8B (10.4\%). Margin drivers: operating margin decreased 2.94pp (26.65\%$\to$23.71\%), driven by SGA ratio $+$1.30pp, R\&D ratio $+$1.22pp, partially offset by tax ratio $-$1.20pp.

Task episodes map 1:1 to ground truth entries (2{,}243~FA tasks total). Each task bundles input references (symbol, period\_end, xbrl\_ref, tools\_available) with a pre-authored natural-language prompt.

\paragraph{Data quality validation.} Each domain pipeline ends with automated validation: PC~(9~checks), RM~(11~checks), FA~(12~checks). Checks include coverage, temporal ordering, point-in-time leakage, cross-source agreement (75.4\% exact match for FA), covariance positive semi-definiteness, and ground truth consistency. All validation reports are stored as JSON.

\subsection{Abbreviated Input/Output Examples by Subtask}
\label{app:examples}

This section shows abbreviated input and expected output for one representative task from each subtask. Full payloads are available in the released dataset.

\paragraph{Unconstrained optimization.}
Input includes objective \texttt{max\_sharpe}, daily expected returns for 30~tickers, a covariance reference, and the risk-free rate. The expected output is a weight dictionary, e.g., AIG 0.060, BF-B 0.060, LNT 0.060, and CPRT 0.048.

\paragraph{Constrained optimization.}
Input adds a constraint object to the optimization inputs: \texttt{long\_only}, maximum weight 0.066, sector cap 0.20, minimum names 18, turnover cap 0.20, and tracking-error cap 0.03. The expected output is a weight dictionary plus a \texttt{constraint\_satisfaction} map, e.g., \texttt{max\_weight}, \texttt{long\_only}, \texttt{min\_names}, and \texttt{max\_names} all true.

\paragraph{Tool-use parameterization.}
Input is a mandate-style natural-language request, such as tracking the benchmark within 3\% tracking error, maximum 5\% per name, long-only, targeting maximum Sharpe ratio. The expected output is an optimizer-call specification with objective \texttt{max\_sharpe}, \texttt{long\_only=true}, \texttt{max\_weight\_per\_name=0.05}, and \texttt{tracking\_error\_limit=0.03}.

\paragraph{Rebalancing.}
Input includes the current portfolio, target objective, covariance reference, and risk-free rate. The expected output includes new weights, a trade list, and total turnover, e.g., AIG increases by 0.011, BAX decreases by 0.039, CMG increases by 0.019, and turnover is 0.15.

\paragraph{Black--Litterman.}
Input contains market-cap weights, analyst views, and a covariance reference. Example views include a relative AAPL--MSFT view with return 0.05 and confidence 0.7, plus an absolute LNT view with return 0.08 and confidence 0.6. The expected output contains posterior returns and optimized weights.

\paragraph{Constraint monitoring.}
Input includes portfolio holdings and mandate constraints such as max position 0.05, max sector 0.25, and beta range 0.8--1.2. The expected output reports overall compliance, violation count, and per-constraint statuses such as AIG failing the position limit, Financial Services failing the sector limit, and beta passing.

\paragraph{Risk identification.}
Input includes holdings plus market-data references for factor exposures and daily returns. The expected output is a ranked risk list, e.g., factor tilt with market beta 0.997, Financial Services sector concentration at 60\%, and AIG single-name exposure at 15\%.

\paragraph{Stress testing.}
Input includes portfolio holdings and a stress scenario such as \texttt{stress\_rates\_down\_100}, with interest-rate and equity-market shocks plus sector sensitivities. The expected output reports portfolio P\&L and attribution, e.g., estimated P\&L of $-0.032$ with sector and factor contribution entries.

\paragraph{Risk remediation.}
Input includes a non-compliant portfolio, a mandate ID, and known violations such as \texttt{sector\_concentration} and \texttt{single\_name\_risk}. The expected output is a trade list and compliance flag, e.g., sell AIG from 0.15 to 0.05, buy ZBRA from 0.00 to 0.03, total turnover 0.25, and post-trade compliance true.

\paragraph{Normalization.}
Input is a ticker-period-form tuple such as TRV, 2024-12-31, 10-K, with \texttt{query\_xbrl} available. The expected output contains canonical metrics such as revenue, operating income, net income, diluted EPS, total assets, total liabilities, stockholders' equity, operating cash flow, operating margin, and EBITDA.

\paragraph{Earnings quality.}
Input is a ticker-period-form tuple such as AAPL, 2022-09-24, 10-Q. The expected output includes Piotroski score and components, Beneish M-score and flag, accruals ratio, income-quality ratio, and flags. Example values include Piotroski 8, Beneish $-2.31$, accruals ratio $-0.0097$, and no flags.

\paragraph{Driver decomposition.}
Input is a ticker and two period ends, such as AAPL from 2017-04-01 to 2017-07-01. The expected output includes revenue delta, top revenue drivers, and margin drivers. Example drivers include Greater China and Europe revenue declines, plus operating-margin and SGA-ratio changes.

\section{System Prompts and Task Prompts}
\label{app:prompts}

\subsection{FinSkillBench Harness (Experiment~05)}

\subsubsection{System Prompt}

The system prompt is identical across all conditions and subtasks:

\begin{quote}
\small
You are a financial analysis agent participating in a controlled evaluation benchmark. Your task is to analyze point-in-time financial data and produce structured outputs. You have access to tools for data retrieval and computation. Use them when needed.

\textbf{Rules:} Use only task-provided data; do not use knowledge after the \texttt{as\_of\_date}. You may call tools to retrieve data and run computations. If skills are available, load a matching skill for guidance; otherwise proceed directly. Submit the final answer with \texttt{submit\_answer} as JSON matching the expected output schema. Do not answer in plain text, do not refuse the task, and use null for fields that cannot be determined.
\end{quote}

\subsubsection{Task Prompt Template}

Each task prompt is constructed programmatically from the episode JSON. For PC and RM subtasks, the template contains the subtask name, \texttt{as\_of\_date}, difficulty, task description, output instructions, compacted input JSON, and an expected-output schema derived from the ground-truth object.

For FA subtasks, the \texttt{task\_description} and \texttt{output\_instructions} are pre-authored natural-language prompts stored in the episode JSON's \texttt{prompt} field. For PC and RM subtasks, these are generated programmatically from the input fields.

The \texttt{compact\_input\_json} applies lossy compaction: covariance matrices are replaced with shape summaries (e.g., ``30$\times$30, full matrix available via tools''), floats are rounded to 3~significant figures, and the total is capped at 12K~characters. The \texttt{schema\_derived\_from\_expected\_output} replaces ground-truth values with type hints (e.g., \texttt{<number>}, \texttt{<boolean>}).

\subsubsection{Condition-Specific Modifications}

\paragraph{No-skill.} The skill-access tools (\texttt{load\_skill}, \texttt{load\_references}) are present in the tool schema but the skill corpus is empty: discovery calls resolve to ``No skills available.'' Domain-specific tools (optimizer scripts, risk calculators) are \emph{not} available. The agent has only \texttt{get\_task\_data}, \texttt{submit\_answer}, and the global \texttt{query\_xbrl} tool (for FA tasks).

\paragraph{Curated.} Human-authored skill packages are mounted. The \texttt{load\_skill} tool description lists available skills (e.g., ``portfolio-construction: Construct optimal portfolio allocations\ldots''). The agent can discover and load SKILL.md content, then call \texttt{run\_skill\_script} to execute validated Python scripts within the skill directory. Domain tools are accessed exclusively through these scripts.

\paragraph{Self-generated.} Curated skill material is withheld. The agent receives the \texttt{save\_skill} tool and an appended instruction:

\begin{quote}
\small
\textbf{Important: Generate Skills First.} Before solving the task, call \texttt{save\_skill} one to three times to capture reusable domain knowledge, formulas, workflows, data-access patterns, or edge cases. Then call \texttt{load\_skill} to read the generated skills back and solve the task using them.
\end{quote}

The agent writes SKILL.md files to a per-episode scratch directory, then reloads them. The skill index is rebuilt after each \texttt{save\_skill} call so subsequent \texttt{load\_skill} calls can find the newly created skills.

\subsection{Hermes Agent Harness (Cross-Harness Validation)}

The Hermes harness uses a nearly identical system prompt:

\begin{quote}
\small
You are a financial analysis agent participating in a controlled evaluation benchmark. Your task is to analyze point-in-time financial data and produce structured outputs. You have access to deterministic financial tools and should use them when computation is needed.

\textbf{Rules:} Use only task-provided data; do not use knowledge after the \texttt{as\_of\_date}. Call tools when needed, submit the final JSON with \texttt{submit\_answer}, do not respond in plain text, use null for unknown fields, and do not refuse the task.
\end{quote}

The key difference is that the Hermes harness uses a \emph{Pattern~B} architecture: heavy data (returns, covariance matrices, factor exposures) is published via a context variable and injected into tool calls server-side, rather than being echoed through model-generated JSON. The user prompt is domain- and subtask-aware, constructed by a dedicated \breakcode{build\_user\_prompt()} function that formats portfolio holdings, mandate constraints, stress scenarios, and analyst views into structured sections.

For the curated condition, Hermes's native \texttt{skills} toolset is enabled (\texttt{skills\_list} + \texttt{skill\_view}), providing progressive disclosure of \texttt{SKILL.md} content. The \texttt{skill\_manage} tool is deregistered to prevent the agent from editing skills during evaluation. All other Hermes built-in toolsets (memory, session search, web, browser, etc.) are disabled to maintain the stateless per-task protocol.

\subsection{Tool Schemas}

Table~\ref{tab:tools} summarizes the tools available to the agent across conditions.

\begin{table}[ht]
\centering
\footnotesize
\setlength{\tabcolsep}{3pt}
\renewcommand{\arraystretch}{1.05}
\caption{Tool availability by condition. ``Always'' means available in all three conditions; ``Curated only'' means available only when the curated skill package is mounted; ``Self-gen only'' means available only in the self-generated condition.}
\label{tab:tools}
\begin{tabularx}{\columnwidth}{@{}>{\raggedright\arraybackslash}p{0.26\columnwidth} >{\raggedright\arraybackslash}p{0.17\columnwidth} >{\raggedright\arraybackslash}X@{}}
\toprule
\textbf{Tool} & \textbf{Availability} & \textbf{Description} \\
\midrule
\texttt{submit\_answer} & Always & Submit final JSON answer (session-ending) \\
\texttt{load\_skill} & Always & Load a \texttt{SKILL.md} into context (returns ``not found'' when empty) \\
\texttt{load\_references} & Always & Load supplementary reference documents for a loaded skill \\
\texttt{get\_task\_data} & Always & Retrieve specific fields from the task input at full precision \\
\texttt{query\_xbrl} & Always (FA) & Query the XBRL financial data panel by ticker, period, and metrics \\
\begin{tabular}[t]{@{}l@{}}
\texttt{run\_skill\_}\\
\texttt{script}
\end{tabular}
& Curated only & Execute a validated Python script from the skill directory \\
\texttt{save\_skill} & Self-gen only & Write a \texttt{SKILL.md} to the per-episode scratch directory \\
\bottomrule
\end{tabularx}
\end{table}

\section{Detailed Subtask Descriptions}
\label{app:subtasks}

This section describes each of the 12~subtasks in detail: what the agent receives as input, what it must produce as output, why the task is difficult, and what cognitive or procedural capabilities it tests.

\subsection{Portfolio Construction}

\subsubsection{Unconstrained Optimization (40 tasks)}
\label{app:sub:uncon}

\paragraph{Input.} A dictionary of expected returns per ticker (30~daily return estimates), a reference to a 30$\times$30 Ledoit--Wolf covariance matrix, a risk-free rate, an objective (\texttt{max\_sharpe}, \texttt{min\_variance}, \texttt{max\_return}, or \texttt{risk\_parity}), and a risk aversion parameter.

\paragraph{Output.} A weight dictionary (\texttt{\{ticker: float\}}) summing to approximately~1.0.

\paragraph{Difficulty.} The agent must solve a convex optimization problem. Without the curated skill (which provides a validated \texttt{optimize.py} script), the agent must either implement mean-variance optimization from scratch within the turn budget or reason about the solution analytically. The covariance matrix contains 900~floating-point values; passing it through model-generated JSON introduces precision loss that changes the optimum. For \texttt{max\_return}, the unconstrained solution is a single-stock portfolio (100\% in the highest-return ticker), which is mathematically correct but counterintuitive.

\paragraph{Cognitive demands.} Numerical optimization, matrix algebra, understanding of mean-variance theory \citep{markowitz1952}, tool parameterization, precision management.

\paragraph{Scorer.} \texttt{l2\_distance\_and\_objective}: $\text{score} = \max(0, \min(1, 1 - L_2 / (4 \times \theta)))$ where $L_2$ is the Euclidean distance between predicted and expected weight vectors and $\theta = 0.05$.

\subsubsection{Constrained Optimization (40 tasks)}
\label{app:sub:con}

\paragraph{Input.} Same as unconstrained, plus a constraint object with fields such as \texttt{long\_only}, \texttt{max\_weight}, \texttt{min\_weight}, \texttt{sector\_limits}, \texttt{min\_names}, and \texttt{max\_names}, and optionally \texttt{max\_turnover}, \texttt{max\_tracking\_error}, and \texttt{factor\_neutrality}. A benchmark weight dictionary and sector mapping are also provided.

\paragraph{Output.} A weight dictionary plus a \texttt{constraint\_satisfaction} dictionary mapping each constraint name to a boolean pass/fail.

\paragraph{Difficulty.} The agent must correctly translate natural-language and structured constraint specifications into optimizer parameters. Constraint interactions are non-trivial: sector limits combined with position limits and minimum-names requirements can produce solutions far from the unconstrained optimum. The agent must also correctly report which constraints are satisfied---a separate verification step from computing the weights.

\paragraph{Cognitive demands.} Constraint formulation, understanding of portfolio mandate structures, multi-objective reasoning, tool parameterization with nested constraint objects.

\paragraph{Scorer.} \texttt{constraint\_satisfaction\_and\_objective}: exact match on all expected \texttt{constraint\_satisfaction} keys acts as a gate. Only if the gate passes is weight $L_2$ scored; missing or incorrect constraint map $\to$ score~0.

\subsubsection{Tool-Use Parameterization (40 tasks)}
\label{app:sub:tool}

\paragraph{Input.} A natural-language scenario description (e.g., ``Build a diversified equity portfolio tracking the benchmark within 3\% tracking error, maximum 5\% per name, long-only''), tool documentation for the optimizer API, and references to covariance and return data.

\paragraph{Output.} A structured \texttt{optimizer\_call} object with the correct \texttt{objective} string and nested \texttt{constraints} dictionary matching the scenario description.

\paragraph{Difficulty.} This subtask isolates the \emph{translation} step: converting a mandate description into correct API parameters. The agent does not need to run the optimizer---it only needs to produce the correct call specification. Errors include wrong objective selection, missing constraints, incorrect numeric thresholds, and wrong parameter names. The task tests whether the agent can read API documentation and map domain concepts to tool interfaces.

\paragraph{Cognitive demands.} Natural language comprehension, API contract understanding, parameter mapping, attention to numeric detail.

\paragraph{Scorer.} \texttt{parameter\_match}: recursively compares the union of predicted and expected parameter leaves, so missing and extra leaves are both penalized. Numeric values match if relative error $< 20\%$; categorical values require exact match. Score = fraction of matching leaves.

\subsubsection{Rebalancing (40 tasks)}
\label{app:sub:rebal}

\paragraph{Input.} A current portfolio (weight dictionary from a previous optimization), a current portfolio date, a covariance matrix reference, a risk-free rate, and an objective.

\paragraph{Output.} A \texttt{trade\_list} (signed weight changes per ticker), \texttt{new\_weights} (post-rebalance weights), and \texttt{turnover} (total one-way turnover).

\paragraph{Difficulty.} The agent must compute optimal new weights \emph{and} derive the trade list and turnover from the difference between current and new weights. Turnover calculation has a specific convention (one-way: $\sum |w_{\text{new}} - w_{\text{old}}| / 2$). The trade list must be consistent with the weight change---a common failure mode is producing trades that do not sum to the correct weight delta.

\paragraph{Cognitive demands.} Portfolio rebalancing mechanics, trade arithmetic, turnover conventions, multi-output consistency.

\paragraph{Scorer.} \texttt{turnover\_compliance\_and\_objective}: weighted combination of post-rebalance weight $L_2$ score (50\%), turnover relative-error ramp (25\%), and trade-list magnitude consistency over expected tickers (25\%).

\subsubsection{Black--Litterman (40 tasks)}
\label{app:sub:bl}

\paragraph{Input.} Market-cap weights (equilibrium prior), a covariance matrix reference, a risk-free rate, and 2--4 analyst views. Each view specifies a type (\texttt{absolute} or \texttt{relative}), symbols, expected return, and confidence level.

\paragraph{Output.} \texttt{posterior\_returns} (per-ticker posterior expected returns) and \texttt{optimal\_weights} (weights from optimizing on the posterior).

\paragraph{Difficulty.} The Black--Litterman model \citep{bl1992} requires: (i)~computing equilibrium returns from market-cap weights and the covariance matrix, (ii)~constructing the pick matrix $P$ and view vector $Q$ from the analyst views, (iii)~computing the posterior via the BL formula involving $\tau$, $\Sigma$, $P$, $Q$, and $\Omega$, and (iv)~optimizing on the posterior returns. Each step can introduce errors. A common failure is using \texttt{expected\_returns} as the prior instead of deriving equilibrium returns from market-cap weights. The model is sensitive to $\tau$ and confidence parameters.

\paragraph{Cognitive demands.} Bayesian updating, matrix algebra, understanding of equilibrium models, multi-step financial computation, correct handling of relative vs.\ absolute views.

\paragraph{Scorer.} \texttt{view\_specification\_and\_weights}: averages (i)~posterior returns MAE score ($\max(0, 1 - \text{MAE}/0.05)$) and (ii)~posterior weights $L_2$ score ($\max(0, \min(1, 1 - L_2/(4 \times 0.10)))$).

\subsection{Risk Management}

\subsubsection{Constraint Monitoring (40 tasks)}
\label{app:sub:cm}

\paragraph{Input.} A portfolio object (holdings with symbols and weights, total NAV), a mandate specification (position limits, sector limits, beta range, long-only, turnover limits), and market data references.

\paragraph{Output.} \texttt{overall\_compliant} (boolean), \texttt{violations\_count} (integer), and a \texttt{constraints} array where each entry has a \texttt{type}, \texttt{status} (PASS/FAIL), worst offender, actual value, and limit.

\paragraph{Difficulty.} The agent must evaluate 4--6 heterogeneous constraints against the portfolio. Constraint types use different units (weights, percentages, ratios, booleans) and require different computations (max over positions, sum over sectors, dot product for beta). The agent must also correctly classify the overall compliance status and count violations. Type normalization is non-trivial: the same constraint may be labeled \texttt{position\_limit\_max}, \texttt{max\_position}, or \texttt{max\_weight} across different mandates.

\paragraph{Cognitive demands.} Mandate interpretation, multi-constraint evaluation, portfolio analytics, attention to edge cases (e.g., beta range is a two-sided constraint).

\paragraph{Scorer.} \texttt{exact\_match}: $0.4 \times \mathbb{1}[\text{compliant match}] + 0.6 \times \text{constraint accuracy}$, where constraint accuracy is the fraction of expected constraint types whose predicted status matches after type and status normalization. The stricter \texttt{jb} scorer variant also withholds the compliance credit unless a constraint list is supplied and penalizes contradictory duplicate predicted constraint types.

\subsubsection{Risk Identification (40 tasks)}
\label{app:sub:ri}

\paragraph{Input.} A portfolio object and market data references (factor exposures, daily returns).

\paragraph{Output.} An array of up to 5 ranked risk exposures, each with \texttt{rank}, \texttt{type} (e.g., \texttt{factor\_tilt}, \texttt{sector\_concentration}, \texttt{single\_name}, \texttt{concentration}), \texttt{detail} (specific description), and \texttt{magnitude}.

\paragraph{Difficulty.} This is the most open-ended RM subtask. The agent must: (i)~compute concentration metrics (HHI, top-$N$ weight share), (ii)~analyze sector distributions, (iii)~review factor exposures for notable tilts, (iv)~synthesize these into a ranked list ordered by magnitude. The ground truth is derived from a deterministic risk engine, but the agent must independently identify and rank the same risks. Without the curated skill's \texttt{compute\_risk\_report.py} script, the agent must perform all computations from raw data. The no-skill baseline is near zero (0.023) because most agents cannot independently compute factor exposures and concentration metrics within the turn budget.

\paragraph{Cognitive demands.} Portfolio risk analytics, factor model interpretation, concentration analysis, multi-criteria ranking, domain knowledge of risk taxonomy.

\paragraph{Scorer.} \texttt{ranked\_list\_recall}: $0.4 \times \text{recall}@k + 0.3 \times \text{precision}@k + 0.3 \times \text{NDCG}@k$, where recall and precision use multiset matching on risk types, and NDCG uses graded relevance from expected magnitudes. $k=5$. In the \texttt{exp05} scorer used by the all-fin Hermes runs, NDCG maps each risk type to the maximum expected magnitude among the top-$k$ expected items and does not clamp the final composite; duplicate predicted types can therefore make this scorer slightly exceed~1. The stricter \texttt{jb} scorer variant uses greedy one-to-one duplicate matching and reports DCG/IDCG diagnostics.

\subsubsection{Stress Testing (40 tasks)}
\label{app:sub:st}

\paragraph{Input.} A portfolio object, a stress scenario specification (scenario ID, name, description, market shocks as factor-level percentage changes), and factor exposure references.

\paragraph{Output.} \texttt{estimated\_pnl\_pct} (decimal portfolio return, e.g., $-0.167$ for $-16.7\%$) and \texttt{attribution} with \texttt{by\_sector} and \texttt{by\_factor} arrays.

\paragraph{Difficulty.} The agent must: (i)~resolve the stress scenario (which may be a historical replay or hypothetical), (ii)~apply factor-level shocks to the portfolio's factor exposures to estimate P\&L, (iii)~decompose the P\&L into sector and factor contributions. A critical failure mode is unit confusion: the scenario specifies shocks as percentages (e.g., ``equity $-20\%$''), but the P\&L must be reported as a decimal return. Another failure mode is attempting to estimate P\&L from headline shock percentages rather than using the full factor model---this produces systematically wrong answers because it ignores cross-factor correlations and idiosyncratic risk.

\paragraph{Cognitive demands.} Factor-based stress testing, P\&L attribution, unit conversion discipline, understanding of historical replay vs.\ hypothetical scenarios.

\paragraph{Scorer.} \texttt{absolute\_error}: $s_{\text{pnl}} = \max(0, \min(1, 1 - (|\hat{y} - y| - 0.1\tau)/(3\tau)))$ where $\tau$ is the P\&L tolerance (typically 1\%). When expected sector attribution is scored, the final score is $0.7\,s_{\text{pnl}} + 0.3\,s_{\text{attr}}$. In \texttt{exp05}, $s_{\text{attr}}$ is the mean per-expected-sector relative-error ramp with a default 5\% relative tolerance; missing expected sectors receive zero and extra predicted sectors are ignored. In the stricter \texttt{jb} scorer, sector attribution is enabled when \texttt{sector\_attribution\_mae} appears in \texttt{verification.metrics}, and $s_{\text{attr}}$ is a triangular ramp on absolute MAE over the union of predicted and expected sectors.

\subsubsection{Risk Remediation (40 tasks)}
\label{app:sub:rr}

\paragraph{Input.} A non-compliant portfolio, a mandate specification, a list of known violations (e.g., ``sector\_concentration'', ``single\_name\_risk''), and market data references including transaction costs.

\paragraph{Output.} A \texttt{trades} array (each with symbol, action, current weight, target weight, trade percentage), \texttt{total\_turnover}, and \texttt{post\_trade\_compliant} (boolean).

\paragraph{Difficulty.} The agent must generate a feasible set of trades that restores compliance while minimizing turnover. This requires: (i)~understanding which constraints are violated and by how much, (ii)~determining which positions to reduce and which to increase, (iii)~ensuring the post-trade portfolio satisfies all constraints simultaneously, (iv)~considering diversification benefits of adding new positions from the broader universe. The curated skill provides a constrained optimizer (\texttt{remediate\_portfolio.py}) that solves this as a minimum-turnover optimization; without it, the agent must reason about trade feasibility manually.

\paragraph{Cognitive demands.} Constrained portfolio optimization, trade generation, multi-constraint satisfaction, turnover minimization, understanding of remediation as an optimization problem.

\paragraph{Scorer.} \texttt{constraint\_satisfaction\_plus\_cost}: $0.3 \times \mathbb{1}[\text{compliant match}] + 0.3 \times \mathbb{1}[\text{turnover OK}] + 0.4 \times F_1(\text{trade pairs})$, where trade pairs are normalized (symbol, action) tuples. The \texttt{exp05} scorer computes a multiset $F_1$; the stricter \texttt{jb} scorer computes set-based precision/recall over normalized trade pairs.

\subsection{Fundamental Analysis}

\subsubsection{Normalization (100 tasks evaluated; 244 total)}
\label{app:sub:norm}

\paragraph{Input.} A ticker symbol, a reporting period end date, a form type (10-Q or 10-K), and access to the \texttt{query\_xbrl} tool.

\paragraph{Output.} A \texttt{metrics} dictionary with 10~canonical financial metrics: revenue, operating\_income, net\_income, eps\_diluted, total\_assets, total\_liabilities, stockholders\_equity, operating\_cash\_flow, operating\_margin, and ebitda.

\paragraph{Difficulty.} The agent must: (i)~query the XBRL panel for the correct ticker and period, (ii)~map returned fields to the expected output schema, (iii)~compute derived metrics that are not directly available (operating\_margin = operating\_income / revenue; ebitda = operating\_income + depreciation\_amortization). The task is relatively straightforward when the \texttt{query\_xbrl} tool is available, which explains the high no-skill baseline (0.681). The curated skill provides no additional benefit ($\Delta = -0.003$) because the global XBRL query tool already provides direct access to the data.

\paragraph{Cognitive demands.} Financial statement literacy, XBRL data navigation, metric computation, understanding of GAAP reporting conventions.

\paragraph{Scorer.} \texttt{metric\_absolute\_error}: unwraps a single \texttt{\{metrics: \{...\}\}} container when both predicted and expected outputs use it. For each scored metric, match if relative error $\leq$ its per-metric tolerance (default 5\%); metadata fields such as \texttt{source}, \texttt{confidence}, \texttt{tier}, and \texttt{xbrl\_fmp\_agreement} are ignored. Score = fraction of scored metrics that match.

\subsubsection{Earnings Quality (100 tasks evaluated; 1{,}002 total)}
\label{app:sub:eq}

\paragraph{Input.} A ticker symbol, a reporting period end date, a form type, and access to the \texttt{query\_xbrl} tool.

\paragraph{Output.} Piotroski F-Score (integer 0--9), 9~binary Piotroski components (e.g., \texttt{roa\_positive}, \texttt{delta\_leverage\_negative}), Beneish M-Score (float), \texttt{beneish\_flag} (boolean: $M > -1.78$), accruals ratio, income quality ratio, and a flags array.

\paragraph{Difficulty.} The agent must: (i)~retrieve financial data for both the current and prior comparable period (same quarter previous year), (ii)~compute 9~Piotroski components requiring year-over-year comparisons of profitability, leverage, and efficiency metrics, (iii)~compute the Beneish M-Score from 8~ratio components (DSRI, GMI, AQI, SGI, DEPI, SGAI, LVGI, TATA), (iv)~compute accruals and income quality ratios, (v)~flag concerns. The multi-period requirement doubles the data retrieval burden. Delta components (e.g., \texttt{delta\_roa\_positive}) require correct identification of the prior period and correct sign interpretation.

\paragraph{Cognitive demands.} Multi-period financial analysis, Piotroski F-Score methodology \citep{piotroski2000}, Beneish M-Score methodology \citep{beneish1999}, ratio computation, temporal reasoning about comparable periods.

\paragraph{Scorer.} \texttt{earnings\_quality\_composite}: renormalized weighted mean over active components: Piotroski component accuracy (episode weight 40\%), Beneish flag match (30\%), and flag $F_1$ (30\%). Numeric ratio accuracy for Beneish M-score, accruals ratio, and income-quality ratio is included only when \texttt{numeric\_ratio\_weight} $> 0$ in \texttt{exp05}/\texttt{zqbok\_experiment05}; the stricter \texttt{jb} scorer defaults that weight to 0.3 if selected.

\subsubsection{Driver Decomposition (100 tasks evaluated; 997 total)}
\label{app:sub:dd}

\paragraph{Input.} A ticker symbol, current and prior period end dates, and access to the \texttt{query\_xbrl} tool.

\paragraph{Output.} \texttt{revenue\_delta} (absolute change in revenue), \texttt{top\_revenue\_drivers} (up to 5 named drivers with type, direction, magnitude in USD, and contribution percentage), and \texttt{margin\_drivers} (operating margin, gross margin, SGA ratio, etc., with direction and delta in percentage points).

\paragraph{Difficulty.} This is the most challenging FA subtask. The agent must: (i)~retrieve aggregate financial data for two periods, (ii)~compute the revenue delta, (iii)~attribute the revenue change to specific business or geographic segments---but the XBRL panel contains only aggregate data, not segment-level data. The curated skill's \texttt{driver\_decomposition.py} script can process segment data when available, but the agent must use domain knowledge about the company's business segments to attribute revenue changes when segment data is unavailable from the tool. (iv)~Compute margin drivers by comparing expense ratios across periods. The ground truth is derived from revenue segmentation data, creating a ceiling effect: agents without access to segment data cannot fully match the expected drivers.

\paragraph{Cognitive demands.} Revenue attribution, segment analysis, margin decomposition, domain knowledge of company business structures, multi-period comparison, understanding of contribution analysis.

\paragraph{Scorer.} \texttt{driver\_f1\_and\_direction}: averages active component scores: revenue-driver name recall/precision (case-insensitive set matching), direction accuracy for matched revenue drivers, numeric magnitude/contribution accuracy for matched revenue drivers when enabled (default enabled), margin-driver recall plus direction accuracy, numeric margin $\Delta$ accuracy when enabled, and revenue-delta relative-error accuracy.

\subsection{Summary of Cognitive Demands}

Table~\ref{tab:cognitive} maps each subtask to the primary cognitive and procedural capabilities it tests.

\begin{table*}[t]
\centering
\footnotesize
\caption{Cognitive and procedural demands by subtask. \checkmark = primary demand; $\circ$ = secondary demand.}
\label{tab:cognitive}
\setlength{\tabcolsep}{3pt}
\renewcommand{\arraystretch}{1.1}
\begin{tabular}{@{}l*{7}{c}@{}}
\toprule
\textbf{Subtask} &
\rotatebox{70}{\textbf{Numerical optim.}} &
\rotatebox{70}{\textbf{Tool param.}} &
\rotatebox{70}{\textbf{Multi-step comp.}} &
\rotatebox{70}{\textbf{Domain knowledge}} &
\rotatebox{70}{\textbf{Data retrieval}} &
\rotatebox{70}{\textbf{Precision mgmt.}} &
\rotatebox{70}{\textbf{Multi-output}} \\
\midrule
Unconstrained optim. & \checkmark & \checkmark & $\circ$ & $\circ$ & $\circ$ & \checkmark & \\
Constrained optim. & \checkmark & \checkmark & $\circ$ & \checkmark & $\circ$ & \checkmark & \checkmark \\
Tool-use param. & & \checkmark & & \checkmark & & & \\
Rebalancing & \checkmark & \checkmark & \checkmark & $\circ$ & $\circ$ & \checkmark & \checkmark \\
Black--Litterman & \checkmark & \checkmark & \checkmark & \checkmark & $\circ$ & \checkmark & \checkmark \\
Constraint monitoring & & & \checkmark & \checkmark & $\circ$ & & \checkmark \\
Risk identification & & & \checkmark & \checkmark & \checkmark & & \checkmark \\
Stress testing & & $\circ$ & \checkmark & \checkmark & \checkmark & \checkmark & \checkmark \\
Risk remediation & \checkmark & $\circ$ & \checkmark & \checkmark & $\circ$ & & \checkmark \\
Normalization & & & $\circ$ & \checkmark & \checkmark & & \\
Earnings quality & & & \checkmark & \checkmark & \checkmark & & \checkmark \\
Driver decomposition & & & \checkmark & \checkmark & \checkmark & & \checkmark \\
\bottomrule
\end{tabular}
\end{table*}

Table~\ref{tab:scorer_formulas} provides the complete mathematical specification of each scorer. The primary formulas match \breakcode{experiments/zqbok\_experiment05/lib/scorers.py} and \breakcode{experiments/jb\_experiment/scoring/scorers\_exp05.py}, which are the \texttt{exp05} scorers used by the all-fin Hermes runs. Where \breakcode{experiments/jb\_experiment/scoring/scorers.py} differs, the stricter \texttt{jb} variant is noted in the subtask descriptions above.

\begin{table*}[ht]
\centering
\small
\caption{Complete scorer specifications. Most component scores are clipped or averaged into $[0, 1]$; the \texttt{exp05} ranked-list scorer does not final-clip NDCG or the composite and can slightly exceed~1 when duplicate predicted risk types receive repeated relevance credit.}
\label{tab:scorer_formulas}
\setlength{\tabcolsep}{5pt}
\renewcommand{\arraystretch}{1.35}
\begin{tabularx}{\textwidth}{@{}>{\raggedright\arraybackslash}p{2.6cm}>{\raggedright\arraybackslash}p{3.6cm}>{\raggedright\arraybackslash}X@{}}
\toprule
\textbf{Scorer} & \textbf{Subtask} & \textbf{Formula} \\
\midrule
L2 distance
  & Unconstrained optim.
  & $\max\!\bigl(0,\;\min\!\bigl(1,\;1 - \lVert w_p - w_e\rVert_2\,/\,(4\theta)\bigr)\bigr)$, \; $\theta{=}0.05$ \\
Constraint gate + L2
  & Constrained optim.
  & All expected \texttt{constraint\_satisfaction} keys must be present and equal; extra predicted keys are ignored. If the gate passes, weight $L_2$ score as above; else~0 \\
Parameter match
  & Tool-use param.
  & Fraction of union leaf parameters matching (numeric: relative error ${<}\,0.2$; categorical: exact; extra/missing leaves count against the denominator) \\
Turnover compliance
  & Rebalancing
  & $0.5\,{\cdot}\,s_{L_2} + 0.25\,{\cdot}\,s_{\text{turn}} + 0.25\,{\cdot}\,s_{\text{trade}}$, where $s_{L_2}{=}\max(0,\min(1,1{-}L_2/0.2))$, $s_{\text{turn}}{=}\max(0,1{-}\min(1,|\hat{T}{-}T|/\max(|T|,10^{-9})))$, and $s_{\text{trade}}$ is mean expected-ticker trade consistency \\
View + weights
  & Black--Litterman
  & Mean of active components: posterior-returns MAE score $\max(0,\,1{-}\text{MAE}/0.05)$ and weight $L_2$ score $\max(0,\min(1,1{-}L_2/(4\theta)))$ with $\theta{=}0.10$ \\
Exact match
  & Constraint monitoring
  & $0.4\,{\cdot}\,\mathbb{1}[\text{compliant match}] + 0.6\,{\cdot}\,\text{constraint accuracy}$, where accuracy is status match over expected normalized constraint types \\
Ranked recall
  & Risk identification
  & $0.4\,{\cdot}\,\text{recall}@k + 0.3\,{\cdot}\,\text{precision}@k + 0.3\,{\cdot}\,\text{NDCG}@k$; recall/precision use multiset type overlap; \texttt{exp05} NDCG uses max magnitude per expected type and is not final-clipped \\
Absolute error
  & Stress testing
  & P\&L: $s_{\text{pnl}}{=}\max\!\bigl(0,\,\min\!\bigl(1,\,1{-}(|e|{-}0.1\tau)/(3\tau)\bigr)\bigr)$; with scored sector attribution, $0.7\,{\cdot}\,s_{\text{pnl}} + 0.3\,{\cdot}\,s_{\text{attr}}$; else $s_{\text{pnl}}$ alone \\
Cost + compliance
  & Risk remediation
  & $0.3\,{\cdot}\,\mathbb{1}[\text{compliant match}] + 0.3\,{\cdot}\,\mathbb{1}[\text{turn.\ OK}] + 0.4\,{\cdot}\,F_1(\text{normalized trade pairs})$ \\
Metric MAE
  & Normalization
  & Fraction of scored non-metadata metrics with relative error ${\leq}$ per-metric tolerance (default 5\%); unwraps \texttt{metrics} wrapper when present on both sides \\
EQ composite
  & Earnings quality
  & Weighted mean over active terms: component accuracy (episode weight 0.4), Beneish-flag match (0.3), flag $F_1$ (0.3); numeric-ratio accuracy is included only if \texttt{numeric\_ratio\_weight}${>}0$ in \texttt{exp05}/\breakcode{zqbok\_experiment05} \\
Driver $F_1$
  & Driver decomposition
  & Mean of active terms: revenue-driver recall/precision, matched revenue-driver direction, matched revenue numeric accuracy, margin recall/direction, matched margin numeric accuracy, and revenue-$\Delta$ accuracy \\
\bottomrule
\end{tabularx}
\end{table*}
\endgroup

\end{document}